# Vehicle-Scene Interaction: A Text-Driven 3D Lidar Place Recognition Method for Autonomous Driving

Tianyi Shang, Zhenyu Li*, Pengjie Xu, Zhaojun Deng

*Abstract*—**Environment description-based vehicle-scene interaction for vehicle localization in large-scale point cloud maps constructed through multi-sensor systems is critically significant for the advancement of large-scale intelligent transportation systems, such as delivery vehicles operating in the 'last mile'. However, current vehicle-scene interaction encounter challenges due to the inability of point cloud encoders to effectively capture local details and long-range spatial relationships and a significant modality gap between text and point cloud representations. To address these challenges, we present Des4Pos, a novel two-stage text-driven 3D Lidar place recognition framework. In the coarse stage, the point-cloud encoder utilizes the Multi-scale Fusion Attention Mechanism (MFAM) to enhance local geometric features, followed by a bidirectional Long Short-Term Memory (LSTM) module to strengthen global spatial relationships. Concurrently, the Stepped Text Encoder (STE)integrates cross-modal prior knowledge from CLIP [1], [2], [3] and aligns text and point-cloud features using this prior knowledge, effectively bridging modality discrepancies. In the fine stage, we introduce a Cascaded Residual Attention (CRA) module to fuse cross-modal features and predict relative localization offsets, thereby achieving greater localization precision. Experiments on the KITTI360Pose test set demonstrate that Des4Pos achieves state-of-the-art performance in text-to-point-cloud place recognition at the expense of a slight increase in the runtime. Specifically, it attains a top-1 accuracy of 40% and a top-10 accuracy of 77% under a 5-meter radius threshold, surpassing the best previous sota method Text2Loc [4] by 8% and 7%, respectively.**

## I. Introduction

NEXT-generation urban ecosystems and intelligent transportation platforms used in urban environments—including self-driving cars, unmanned aerial vehicles, and automated delivery networks—require precise self-localization through text-driven vehicle-scene interaction. Conventional visual place recognition (VPR) methods [5], [6], [7], [8], [9], [10] primarily use monocular imagingto establish location correspondences. LiDAR-based Place Recognition (LPR) [11], [12] uses LiDAR sensing to establish location correspondences. The core idea behind both VPR and LPR is to obtain a global descriptor for a point cloud or image. This descriptor is then matched for similarity against a database to determine the final location coordinates. In addition to surveys on VPR [13], [14] and LPR [15], [16], some recent surveys [17] have also incorporated discussions on Cross-Modal Place Recognition.

Nevertheless, two critical limitations are observed in VPR and LPR: suboptimal human-machine collaborative efficiency and compromised robustness when confronted with temporal environmental shifts or observational perspective alterations. The emerging paradigm of text-to-point-cloud localization addresses these limitations by establishing direct correlations between textual descriptions and point-cloud spatial signatures, while also eliminating the dependency on physical proximity constraints during place recognition.

The text-guided point-cloud localization task presents two challenges: (1) Semantically vague textual queries often generate ambiguous mappings to multiple candidate regions in a large-scale point-cloud map, and (2) Descriptions of adjacent geographic areas exhibit minimal linguistic distinctions, posing significant challenges for high-precision coordinate prediction. To address these challenges, Text2Pos [18] proposed a novel coarse-to-fine architecture as follows: during the coarse localization stage, large-scale point-cloud maps are divided into spatial submaps for text-submap alignment, while the fine stage fuses multimodal embeddings to regress coordinates within each candidate submap. Wang et al. [19] proposed the Relation-Enhanced Transformer (RET) to address the limitations of Text2Pos in modeling instance-level relations. Text2Loc [4], enhances performance in the coarse stage through contrastive learning while significantly reducing computational costs in the fine stage using a matching-free mechanism. The latest work, MNCL [20], improves its performance by designing a novel loss function.

To overcome these challenges, we develop a two-stage vehicle place recognition framework for autonomous driving-Des4Pos. Previous methods such as Text2Loc used the same encoder to simultaneously encode point clouds and text, while Des4Pos employed different encoders, aiming to bridge the modality differences.

To construct discriminative global descriptors for point clouds in the coarse stage, we propose a novel Multi-scale Fusion Attention Mechanism (MFAM). MFAM first performs self-attention computation across multiple scales and subsequently employs an attention-weight-based fusion strategy to integrate these multi-scale features, ultimately generating point-cloud descriptors incorporating fine-grained local geometric details (the details of individual objects and the spatial relationships between neighboring objects). To further enhance long-range spatial dependencies, we integrate a bidirectional LSTM, leveraging its sequential memory gate mechanism to effectively capture spatial dependencies between distant key regions within the point-cloud map.

Current methods primarily focus on modality fusion while overlooking the essential role of cross-modal disparity in offset prediction during the fine-tuning stage. We introduce the Cascaded Residual Attention (CRA) module to address this limitation. The CRA module facilitates deep interaction between text and point-cloud features through a series of cascaded cross-attention mechanisms. Additionally, a residual-skip connection is incorporated from the query stream to the feature map at each cross-attention layer. These residual connections preserve the original modality-specific features, which inherently capture spatial misalignment information between text and point clouds, thereby enhancing the accuracy of relative positional offset predictions.

We conduct comprehensive experiments on the KITTI360Pose [21] dataset for Des4Pos. The results demonstrate that our method significantly outperforms previous state-of-the-art approaches, and extensive ablation studies further validate the impact of each module on accuracy. Our main contributions can be summarized as follows:

- We propose MFAM in conjunction with a bidirectional LSTM for point cloud encoding. The MFAM integrates multi-scale self-attention and cross-attention-based feature fusion to enhance local geometric representation. Meanwhile, the bidirectional LSTM leverages sequential modeling to encode spatial dependencies between distant regions, thereby improving descriptor distinctiveness.
- We design a Stepped Text Encoder (STE) to bridge the cross-modal gap between text and point clouds. The STE first learns cross-modal prior knowledge from CLIP and subsequently projects descriptions from the text-image joint latent space into the text-to-point-clouds joint latent space, thereby facilitating robust cross-modal alignment.
- We introduce a CRA module for disparity-aware cross-modal fusion. Through cascaded cross-attention layers with residual-skip connections, CRA fuses text and point-cloud features while preserving cross-modal disparity characteristics, thereby enabling precise prediction of relative positional offsets.

## II. Related Work

### A. LiDAR Place Recognition for Autonomous Driving

LiDAR-based place recognition (LPR) methods primarily focus on global descriptor retrieval by processing point clouds or their Bird's Eye View (BEV) projections. In the early evolution of LPR, PointNetVLAD [22] leverages the strengths of PointNet [23] and NetVLAD [24] to generate global descriptors. BEV-based approaches, such as BevPlace [11], enhance rotational invariance through group convolutions. Additionally, I2P-Rec [25] extracts consistent descriptors from images and point clouds by projecting them into BEV for place recognition. [26] presents a novel localization framework for autonomous driving with visual-inertial-liDAR fusion in underground space.

Hybrid point-based LPR frameworks utilize attention graphs or 3D-CNNs [11] on pose graphs to extract structural features from point clouds, improving place recognition performance. For example, LPD-Net [27] captures local feature distributions through graph-based aggregation, while DH3D [28] and SOE-Net [29] enhance 6-DoF pose estimation by using local feature extractors and PointOE modules. Additionally, InCloud [30] addresses the challenge of incremental learning in LPR by preserving higher-order embedding spaces.

Among transformer-based LPR methods, TransLoc3D [31] leverages adaptive attention mechanisms to capture global contextual information, NDT-Transformer [32] extracts global descriptors from point-clouds compressed by the normal distribution transform, and PPT-Net [12] captures multi-scale 3D features through pyramid transformers.

Sparse 3D-convolution-based LPR methods, such as MinkLoc3D [33], demonstrate superior performance compared to transformer-based methods. Recent advancements include LCDNet [34], which integrates unbalanced optimal transport theory for point-clouds alignments. Meanwhile, EgoNN [35] employs a fully convolutional architecture to generate a global representation from sparse voxelized data. Additionally, LoGG3D-Net [36] proposes a novel local consistency loss, which enhances LPR performance. [37] proposes a sparse LiDAR PCC system to compress the point cloud geometry, which contains a back channel allowing the decoder to initiate such a request to the encoder.

### B. Cross-modal Place Recognition for Autonomous Driving

Language-driven 3D place recognition achieves coordinate localization within 3D point clouds using natural language queries presents a two-stream network, which decomposes video and text alignment into multiple dimensions. Text2Pos [18] is the first to propose a coarse-to-fine framework for this task and establishes the KITTI360Pose dataset. The coarse stage identifies candidate regions by projecting text embeddings and point-cloud features into a cross-modal aligned latent space. The fine stage subsequently refines coordinates through the accumulation of offsets within candidate regions.

In the KITTI360Pose dataset, textual descriptions provide hints as separate elements. However, Text2Pos fails to explicitly model the connections between these hints, thereby hindering the learning of instance-level relationships. RET [19] addresses this shortcoming by utilizing self-attention modules to capture intra-modal hint relationships in the coarse stage and cross-attention modules to enhance multi-modal feature fusion during the fine stage.

Building upon these foundations, Text2Loc [4] attains state-of-the-art performance through the integration of pre-trained T5 language models [38] with transformer architectures. Its matching-free strategies in the fine stage employ cascaded cross-attention transformers to optimize multi-modal fusion, while contrastive learning strategies in the coarse stage substantially improve localization precision by enhancing feature distinction in the joint embedding space. Liu et al.'s method [20] improves cross-modal place recognition by tackling the ”mismatched factors” between different data types (like point clouds and text). Instead of just training the model to recognize matches, their approach focuses on making mismatched pairs as dissimilar as possible. This is done by minimizing the sim- ilarity of different locations at three levels of detail, including global-level, instance-level and relationlevel, respectively.

Simple point-cloud encoders in current methods often miss key structural and semantic details. Des4Pos overcomes this by combining Multi-scale Fusion Attention Mechanism(MSAM) and bidirectional LSTM to create comprehensive global representations. Des4Pos also facilitates a Stepped Text Encode(STE) to bridge text-to-point-cloud modality discrepancies in the coarse stage. In the fine stage, the integration of skip connections and cross-attention enables cross-modal fusion while preserving original modality-specific features, thereby improving the prediction accuracy of relative positional offsets.

### C. Text-To-point-cloud Understanding

Emerging works focus on bridging natural language and point-cloud data for multimodal comprehension. Among these efforts, ScanRefer [39] proposes a novel task of grounding linguistic descriptions to specific objects in 3D environments. In contrast, ReferIt3D [40] focuses on fine-grained language-guided 3D object localization in indoor scenarios that contain multiple similar objects. Building on this, InstanceRefer [41] first reformulates the 3D visual grounding task as an instance-matching problem and then performs multi-level contextual inference to determine the target object's position. [42] utilizes various graph-based architectures to effectively capture textual and point-cloud features, setting a new benchmark in 3D language grounding.

Notably, PointCLIP [43] extends the CLIP architecture to 3D point-cloud recognition through multi-view depth map projection, achieving groundbreaking progress in text-to-point-cloud understanding.

Recent advancements in multimodal large language models (MLLM) have demonstrated impressive capabilities in processing not only text but also images and point clouds. Current studies have increasingly explored text-to-point-cloud understanding based on MLLM. For instance, Yu et al. [44] propose a tri-modal framework that significantly improves text-to-point-cloud retrieval accuracy through cross-modal alignment of text, image, and point-clouds. Additionally, SegPoint [45] leverages the semantic reasoning capabilities of MLLM to address multiple 3D segmentation tasks in a unified framework. Furthermore, PointLLM [46] integrates specialized point-cloud encoders with MLLM, allowing it to understand point-cloud according to human instructions and generate appropriate responses.

## III. Methodology

The coarse-to-fine Des4Pos framework, as shown in Fig. 1, achieves accurate location regression in large-scale point-cloud maps by leveraging environmental language descriptions.

### A. Problem Statement

Des4Pos represents the point-cloud map as: $M_{ref} = \{m_i : i = 1, ..., M\}$, where each $m_i$ is a cubic submap containing multiple object instances $I_{i,j}$. The textual queries are formalized as a set $T = \{t_i\}_{i=1}^{n}$, where each query $t_i \in T$ is composed of spatial relationship hints: $t_i = \{h_k^{(i)}\}_{k=1}^{h}$. The objective of the coarse stage is to find the top-$k$ most likely submaps $m_i$ given a query $t_i$. To this end, we design a dual-branch encoder $F$ consisting of a point-cloud encoder and a text encoder. It projects the textual query and the point-cloud submaps into a shared semantic space to maximize the semantic similarity between matching pairs.

In the fine stage, Des4Pos aims to precisely predict the location $(x, y)$ within each submap. The localization objective is formulated as follows:

$$\min E \left\| (x, y) - \phi\left( t_i, \underset{m_i \in M_{\text{ref}}}{\operatorname{argmin}} \{ d(F(t_i) F(m_i)) \} \right) \right\|_1^2 . \quad (1)$$

where:

- $t_i$ is the $i$-th description query,
- $m_i$ is the $i$-th point-cloud submap,
- $d(.)$ is the Euclidean distance between the encoded representations of the textual description and the point-cloud submaps,
- $\phi$ is a multi-layer perceptron (MLP) network that predicts the final coordinates $(x, y)$.

The above formulation enables the model to first narrow down the possible locations to the most likely submaps and then refine the coordinates within each submap to predict the final position accurately.

### B. Coarse stage

*1) Point-Clouds Encoder:* Cross-modal alignment requires a comprehensive representation of point-cloud. In the coarse stage, Des4Pos represents the point-cloud map as: $M_{ref} = \{m_i : i = 1, ..., M\}$, where each submap $m_i$ contains multiple object instances.

Inspired by PointNet [23], we then use a permutation-invariant symmetric function to handle the unordered nature of the point-cloud:

$$f(P) = \gamma\left( \max_{m_i \in M_{ref}} \chi(m_i) \right) . \quad (2)$$

where $\chi(.)$ is a nonlinear function and $\gamma(\,)$ is an MLP.

After encoding the point cloud with a PointNet-like structure, we realized that several distinct features are critical for cross-modal matching. These include object color, center position (as positional relationships are key for this task), and the density of points within an object, which implicitly carries information about its category. To enhance these features, we first extract the RGB coordinates, instance center coordinates ($C_{\text{center}}$), and instance density ($S_{\text{density}}$) from the raw point cloud. We then process each of these through dedicated encoders: a color encoder ($c = \text{MLP}_{\text{color}}(\text{RGB})$), a position encoder ($p = \text{MLP}_{\text{pos}}(C_{\text{center}})$), and a density encoder ($r = \text{MLP}_{\text{density}}(S_{\text{density}})$). This density encoder, for instance, provides prior knowledge about object categories by accounting for variations in point density across instances. Finally, we concatenate $f(P)$ with the three encodings (color $c$, position $p$, and intensity $r$) and fuse them to obtain a comprehensive representation of the given point-cloud submap:

$$s_i = \sigma\left(W \text{Concat}\left(f(P), c, p, r\right) + b\right) . \quad (3)$$

Where $\sigma(\cdot)$ is the activate function, $W$ and $b$ represent weights and bias terms of this model. Through the aforementioned feature extraction pipeline, we obtain a set of comprehensive descriptors $S_{submap} = \{s_i : i = 1, ..., S\}$ for submaps $m_i \in M_{ref}$.

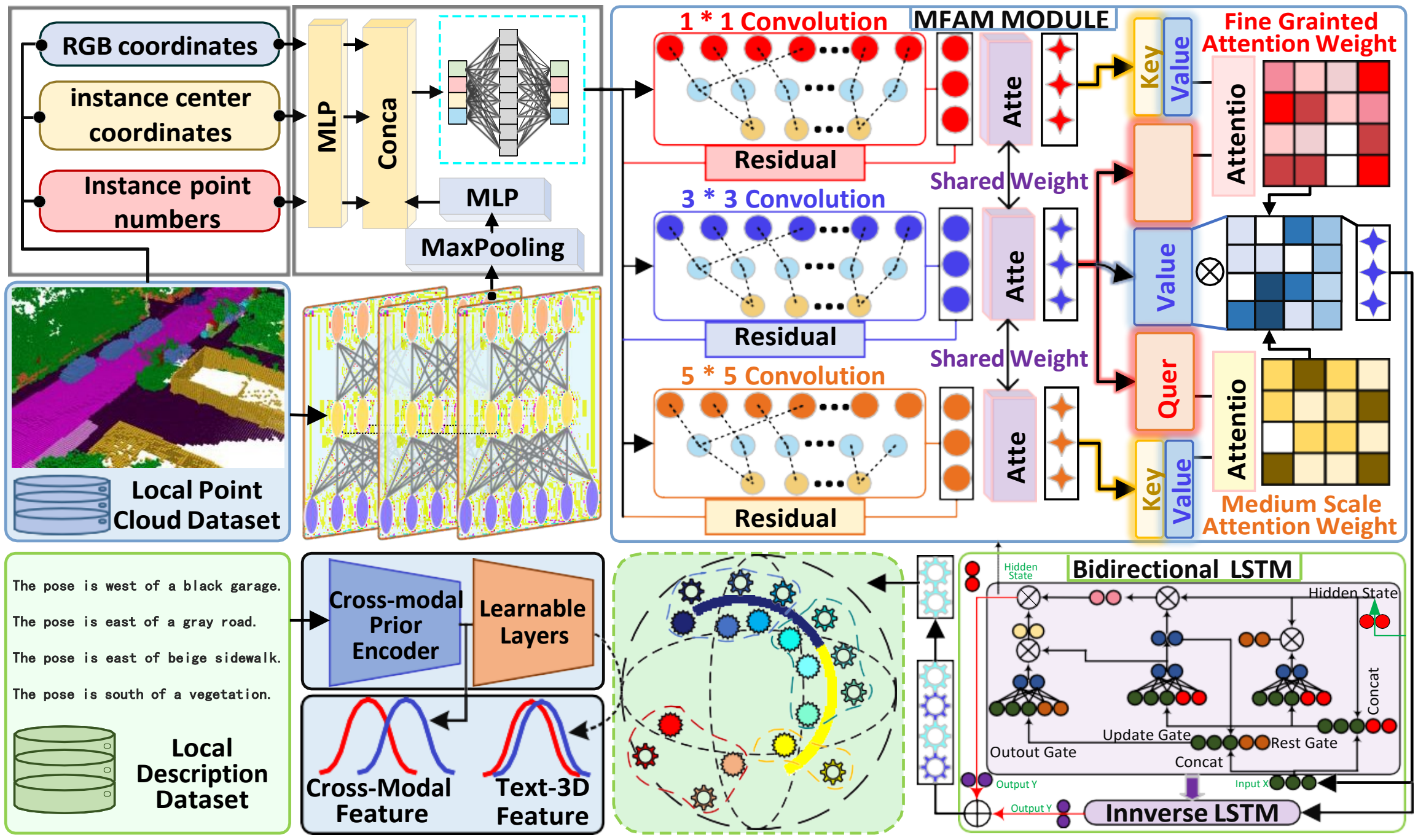

Fig. 1. In the coarse stage, Des4Pos aims to achieve text-to-point-cloud alignment. We first encode color, position, and intensity information separately to comprehensively represent the point cloud. Then, an MSFM is used to enhance local geometry via multi-scale interactions. Finally, a bidirectional LSTM is used to capture global spatial dependencies, generating a discriminative point-cloud descriptor. These point-cloud descriptors and text descriptors are then projected into a unified semantic space via contrastive learning, thereby achieving cross-modal alignment.

After completing the initial features of point-cloud submaps, we focus on generating discriminative global descriptors. To this end, Des4Pos first emphasizes modeling local geometric features, followed by enhancing global spatial relationships within submaps.

To comprehensively capture local characteristics, we introduce a novel Multi-scale Fusion Attention Mechanism (MFAM), as shown in Fig. 1. MFAM first scans the point-cloud submaps using parallel $1\times1$, $3\times3$, and $5\times5$ convolution kernels with residual connections to extract fine-grained detail features $F_1$, medium-scale structural features $F_3$, and wide-range perceptual features $F_5$, respectively. Subsequently, the features at each scale are enhanced by self-attention modules that capture the geometric dependencies between instances. This process is formulated as:

$$F_i = Attention\left(\sigma\left(k_i * s_i + b_i\right) + s_i\right). \tag{4}$$

where $k_1$, $k_3$, and $k_5$ denote the $1\times1$, $3\times3$, and $5\times5$ convolution kernels, respectively.

Subsequently, MFAM employs a novel attention-weight-based strategy to fuse multi-scale local features through three cross-attention operations.

First, cross-attention is performed between $F_3$ and $F_1$ to generate an attention weight R1, which aggregates fine-grained geometric details. Next, the second cross-attention operation is conducted between $F_3$ and $F_5$ to compute an attention weight R2, which fuses information from a moderately expanded receptive field.

Finally, a new attention weight R3 is obtained through element-wise multiplication of R1 and R2, achieving a balanced integration of fine-grained features and moderately expanded receptive field coverage. Using R3 as the attention weight and $F_3$ as the value (V), we perform the third cross-attention operation:

$$d_i = Attention\left(\sigma\left(\frac{F_1 F_3^T}{\sqrt{d_k}}\right) * \sigma\left(\frac{F_5 F_3^T}{\sqrt{d_k}}\right), F_3\right). \tag{5}$$

where $\sigma$ represents the Softmax operation and the first module of *Attention* is the new attention weight map.

The feature map $d_i$ preserves mid-scale structural coherence while capturing fine-grained point-wise details and expanded contextual features, thus effectively representing local geometric structures in point-cloud.

After comprehensively characterizing local features, Des4Pos focuses on modeling spatial correlations at the global scale. To this end, we employ a bidirectional LSTM to process locally enhanced point-cloud submap $d_t$. This process can be mathematically expressed as:

$$LSTM = \begin{cases} f_t = \sigma(W_f \cdot [h_{t-1}, d_t] + b_f) \\ i_t = \sigma(W_i \cdot [h_{t-1}, d_t] + b_i) \\ \tilde{C}_t = \tanh(W_C \cdot [h_{t-1}, d_t] + b_C) \\ C_t = f_t \odot C_{t-1} + i_t \odot \tilde{C}_t \\ o_t = \sigma(W_o \cdot [h_{t-1}, d_t] + b_o) \\ h_t = o_t \odot \tanh(C_t) \end{cases}, \tag{6}$$

$$\{h_t\}_{t=1}^{T} = [\mathrm{LSTM}_{\rightarrow}(d_t, h_{t-1}), \mathrm{LSTM}_{\leftarrow}(d_t, h_{t+1})].$$

where $h_t$ is the hidden state, $C_t$ is the cell state storing long-term memory, and $f_t$ is the forget gate controlling memory retention.

The effectiveness of bidirectional LSTM in Des4Pos manifests in three aspects: Firstly, in text-to-point-cloud cross-modal place recognition tasks, orientation-aware descriptors serve as discriminative features. In the point clouds, this feature is represented as spatial relationships between instances. bidirectional LSTM leverages its sequential architecture to capture extended spatial correlations among point-cloud instances, thereby capturing long-range dependencies in the point-cloud.

Secondly, human attention prioritizes salient entities (e.g., buildings, landmarks) while filtering transient objects (e.g., road surfaces, dynamic objects). The gated memory units in LSTM—particularly forget gates—mimic this selectivity by adaptively attenuating non-salient signals, thus preserving discriminative feature representations for cross-modal alignment.

Thirdly, Des4Pos deploys a bidirectional LSTM with dual temporal processors: forward (past→future) and backward (future←past) layers. This configuration enables each hidden state to fuse bidirectional context, thereby enhancing point-cloud feature descriptor learning.

After processing by the bidirectional LSTM, hidden states $\{h_t\}_{t=1}^{T}$ undergo strided downsampling-based feature compression to generate a global descriptor $n_i$ for the point-cloud submaps. The computational flow can be formally expressed as:

$$n_i = \{h_t \mid t = 1 + (j-1)k,\ j = 1, 2, \ldots, \lfloor T/k \rfloor\} \in R^d. \quad (7)$$

where $k$ is the step size and $T$ denotes the sequence length.

*2) Text Encoder:* In the coarse stage, the primary objective of Des4Pos is to bridge the cross-modal representation gap between text and point cloud. To this end, the Stepped Text Encoder (STE) adopts a two-stage progressive strategy: The first stage transfers cross-modal prior knowledge, followed by text-to-point-cloud feature alignment in the second stage. This staged design effectively reduces modality discrepancies and ultimately enables effective cross-modal representation matching, as shown in Fig. 2.

During the cross-modal prior knowledge transfer stage, STE employs a frozen T5 model as the backbone to leverage its long-text encoding capabilities. To inject cross-modal prior knowledge into text representations, we design a cross-modal prior head at the end of T5, consisting of stacked Transformer layers, MLP, and Max Pooling. In the training phase, short text descriptions $R$ are fed into STE and the CLIP text encoder and the cosine similarity loss is computed between their output features. This process can be formulated as:

$$\mathcal{L} = 1 - \cos\left(f_{\mathrm{STE}}(R), f_{\mathrm{CLIP}}(R)\right). \quad (8)$$

where $f_{\mathrm{STE}}(\cdot)$ represents the encoded output processed through the frozen T5 backbone and the learnable cross-modal prior head, and $f_{CLIP}(\cdot)$ represents the output of the CLIP function.

Notably, directly using CLIP as the backbone would degrade performance due to its limited long-context encoding capability. STE's first stage preserves T5's long-context processing advantages while transferring CLIP's cross-modal prior knowledge to the text encoding process. This approach lays the foundation for downstream text-to-point-cloud alignment.

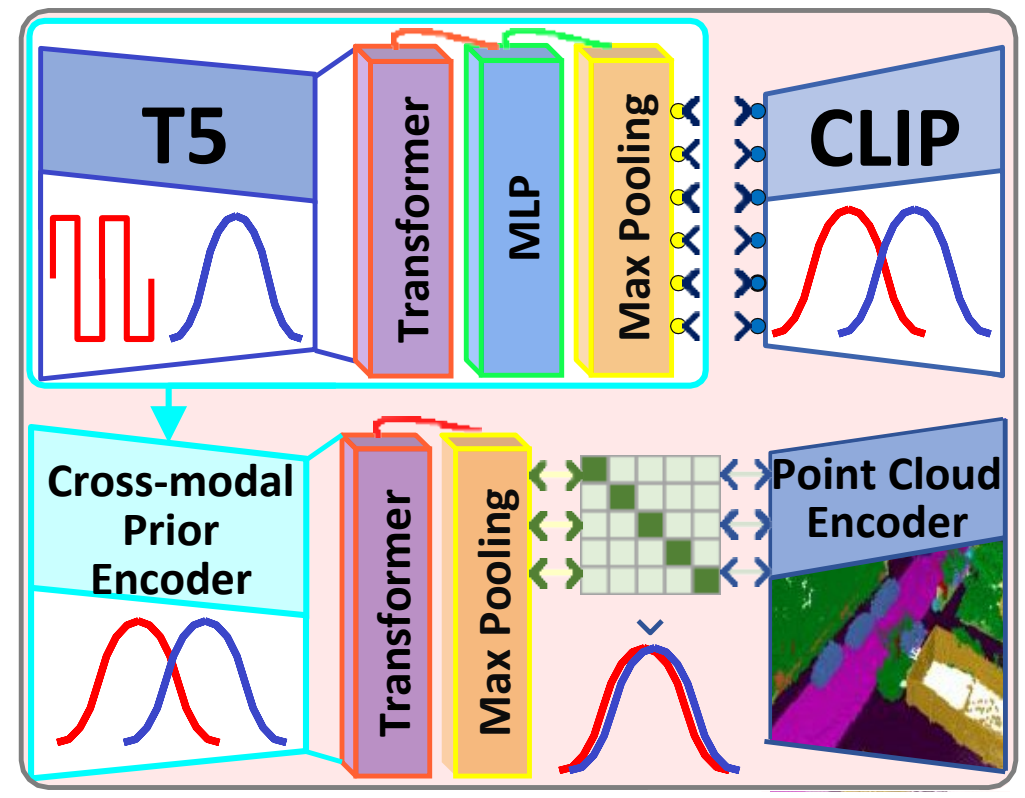

Fig. 2. The Stepped Text Encoder (STE) model employs a progressive encoding strategy for text feature learning. As illustrated in the figure, square wave signals represent textual domain features while sinusoidal waves correspond to cross-modal domain features. During the first stage, we integrate text-to-image cross-modal prior knowledge acquired from the CLIP pre-trained model into the text encoder. In the second stage, STE transfers these cross-modal priors into the text-to-point-cloud feature space, ultimately generating textual feature representations optimized for text-to-point-cloud alignment.

In the text-to-point-cloud feature alignment stage, both the T5 encoder and the cross-modal prior are frozen. A cross-modal alignment head with a Transformer-MLP-Transformer architecture is then added to the frozen modules. The first Transformer captures the word-level semantic representations, while the subsequent MLP projects these features into a high-dimensional latent space. Then, the second Transformer layer establishes inter-sentential semantic coherence, constructing a final global textual descriptor $d_i$. $d_i$ contains both semantic information and cross-modal priors. It is subsequently used to establish text-to-point-cloud alignment through the loss function detailed in following section).

*3) Loss Function:* After obtaining the final encoded point-cloud submap representations $N_{ref} = \{n_i : i = 1, \ldots, N\}$ and text representations $D = \{d_i\}_{i=1}^{n}$, the last stage of the STE aims to map both point-cloud features $n'_i$ and text features $d'_i$ into a unified embedding space. Therefore, we apply contrastive learning to align features: For matched text and point-cloud pairs, embeddings are pulled closer by minimizing the cosine distance. For unmatched pairs, embeddings are pushed apart by maximizing the distance. Given a point-cloud descriptor $n'_i$ and a text descriptor $d'_i$, the contrastive loss can be formulated as:

$$\mathcal{L} = -\log \frac{\exp\left(n'_i \cdot d'_i/\tau\right)}{\sum_{j \in N} \exp\left(n'_i \cdot d'_j/\tau\right)} - \log \frac{\exp\left(d'_i \cdot n'_i/\tau\right)}{\sum_{j \in N} \exp\left(d'_i \cdot n'_j/\tau\right)}. \quad (9)$$

where $\tau$ is the temperature coefficient.

## C. Fine stage

In the coarse stage, Des4Pos performs top-K point-cloud submap retrieval using a text query. In the fine stage, Des4Pos predicts precise 2D coordinates within the point-cloud submap. To achieve this, the textual descriptions and point cloud are first encoded separately, and these two features are then fused and input into a multi-layer perceptron (MLP). The MLP outputs a 2D offset, which is then added to the 2D center point coordinates of the point cloud submap to produce the final predicted coordinates.

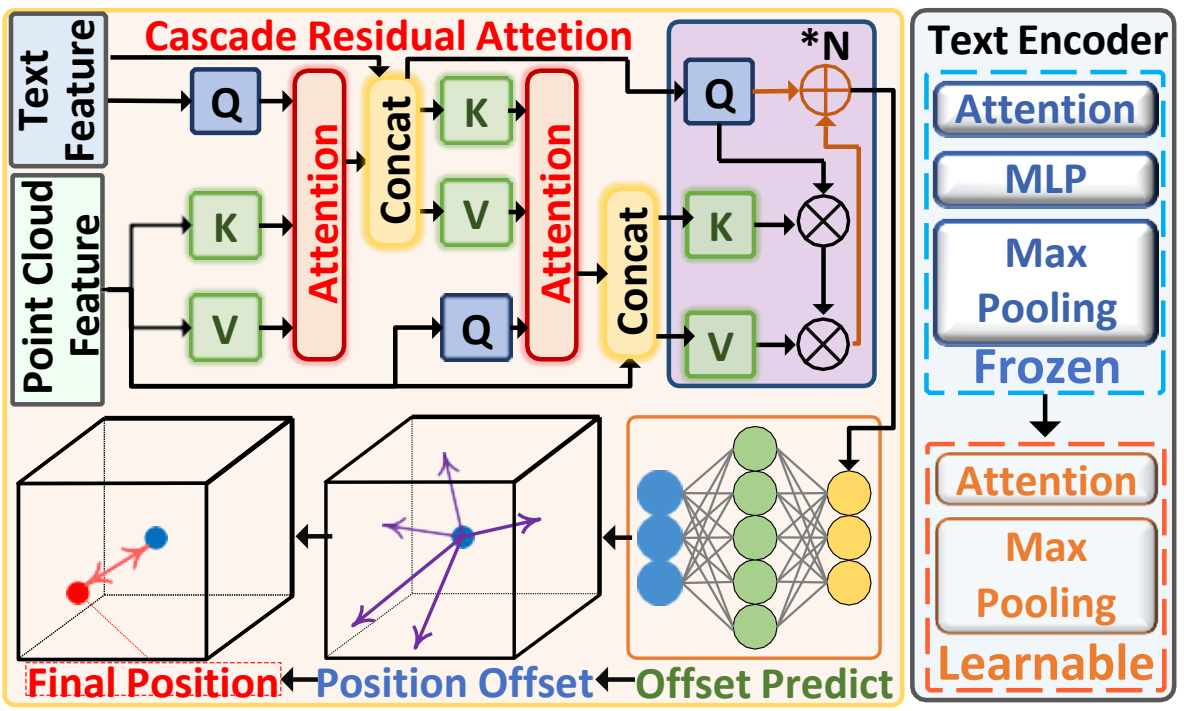

Fig. 3. The fine stage of the Des4Pos aims to predict final coordinates in point-cloud submaps. To this end, we employ Cascaded Residual Attention (CRA) to enable deep fusion of text and point-cloud features. Specifically, CRA enables cross-modal feature fusion through cascaded attention, while preserving feature discrepancies through skip-residual connections. Subsequently, we use an MLP to predict offsets from the submap's center based on CRA's fused features. These offsets are finally mapped to global coordinates. Notably, we predict two-dimensional coordinates for the final position.

To precisely determine target coordinates $(x, y)$ within candidate submaps in the fine stage, we propose the novel Cascaded Residual Attention (CRA) mechanism for cross-modal fusion, as shown in Fig. 3. The fused features are then used to estimate position offsets relative to submap centroids.

During the fine stage, the text encoder combines frozen components (pre-trained T5 and cross-modal prior head) with trainable components (a Transformer layer and a max pooling module). Simultaneously, we directly reuse the point-cloud encoder pre-trained in the coarse stage. The encoded point-cloud embeddings $n_i$ and text embeddings $t_i$ are subsequently fed into the CRA to facilitate feature fusion.

CRA initially generates fused features $R1$ through cross-attention, using text embeddings $t_i$ as query $(Q)$ and point-cloud embeddings $n_i$ as key $(K)$ and value $(V)$. This process incorporates residual connections to preserve original text features, which can be formulated as:

$$R_1 = \text{Residual}\left(\text{Attention}(t_i, n_i, n_i),\ \ T\right). \tag{10}$$

To deepen multimodal fusion, CRA generates fused features $R2$ via cross-attention with residual connections. This cross-attention operation uses point-cloud embeddings $n_i$ as query $(Q)$, $R1$ as key $(K)$ and value $(V)$, which can be formally expressed as:

$$R_2 = \text{Residual}\left(\text{Attention}(n_i, R_1, R_1),\ \ n_i\right). \tag{11}$$

The above process integrates cascading attention with residual connections to produce fused features $R1$ and $R2$. Subsequent fusion layers iteratively refine cross-modal representations using these features. Each fusion step is formally defined as:

$$R_i = \text{Residual}\left(\text{Attention}(R_{i-2}, R_{i-1}, R_{i-1}),\ \ R_{i-2}\right), \quad i = 3, 4, 5, 6, \ldots, n. \tag{12}$$

where $R_i$ represent the fused text-to-point-cloud features, $Attention(\ )$ is the output of attention module.

The CRA employs cascaded cross-attention mechanisms to achieve thorough cross-modal fusion. Then, residual connections are incorporated to preserve the original features of point-cloud and text embeddings. This design effectively maintains cross-modal discrepancy information, which provides critical feature support for offset prediction.

Des4Pos then feeds the cross-modal fused features $R_i$ into an MLP to predict the offsets relative to the point-cloud submap center. Subsequently, Des4Pos predicts the final coordinates by accumulating these offsets. The loss function for the fine stage, which aims to minimize the Euclidean distance between the predicted and ground-truth coordinates, can be formulated as:

$$L\left(C_{\text{gt}}, C_{\text{pred}}\right) = \|C_{\text{gt}} - C_{\text{pred}}\|_2. \tag{13}$$

where $C_{\text{pred}} = (x,\ y)$ represents the predicted target coordinates and $C_{\text{gt}}$ denotes the ground truth coordinates.

## IV. Experiment

### A. Dataset and Implementation Details

The KITTI360Pose dataset spans nine urban zones totaling 15.51 km² and consists of 14,934 geolocated positions with corresponding textual annotations. For experimental purposes, the dataset is divided into three parts: training (5 zones), validation (1 zone), and testing (3 zones). During the coarse stage of Des4Pos, we partition the point-cloud map into 30 m × 30 m submaps using a 10 m stride to establish matching units for text-to-point-cloud retrieval. We then construct a dataset comprising 100,000 short sentence descriptions specifically for training the SET model.

We train Des4Pos on a hardware platform comprising a 128-core AMD EPYC processor and an NVIDIA RTX 4090D GPU (24 GB). The detailed parameter configurations of the coarse-to-fine process are as follows: During the coarse stage, we set the batch size to 64 with a base learning rate of 0.0005 and train for 20 epochs to generate 256-dimensional feature vectors; in the fine stage, we adjust the batch size to 32 with a learning rate of 0.0003 and train for 35 epochs. This configuration ensures comparability with existing methods on the KITTI360Pose dataset.

### B. Performance Analysis

*1) Performance Comparison with SOTA Methods:* We conduct comprehensive experiments on the KITTI360Pose dataset, comparing our proposed Des4Pos with existing SOTA methods. As shown in Table I, compared to the most advanced method Text2Loc, Des4Pos achieves improvements of **8%**, **8%**, and **7%** in the top1/5/10 accuracy within a 5m range on the validation set, and **7%**, **8%**, and **8%** on the test set, respectively. Given that localization accuracy within a 5m range mostly reflects the precise localization capability of the model, these results fully demonstrate the superior performance of Des4Pos.

TABLE I
COMPARISONS OF THE PROPOSED DES4POS WITH EXISTING SOTA METHODS ON THE KITTI360POSE DATASET. MNCL+DES4POS SIGNIFIES THE INTEGRATION OF THE TEXT ENCODER (STE), THE POINT CLOUD ENCODERS (MFAM AND LSTM), AND THE CRA MODULE USED FOR FEATURE FUSION IN THE FINE STAGE OF OUR DES4POS METHOD INTO THE MNCL ARCHITECTURE.

| Methods | Localization Recall ($\epsilon < 5/10/15m$) ↑ | | | | | |
|---|---|---|---|---|---|---|
| | Validation Set | | | Test Set | | |
| | k = 1 | k = 5 | k = 10 | k = 1 | k = 5 | k = 10 |
| NetVLAD [24] | 0.18/0.33/0.43 | 0.29/0.50/0.61 | 0.34/0.59/0.69 | 0.12/0.15/0.17 | 0.22/0.32/0.34 | 0.24/0.29/0.31 |
| PointNetVLAD [22] | 0.21/0.28/0.30 | 0.44/0.58/0.61 | 0.54/0.71/0.74 | 0.13/0.17/0.18 | 0.28/0.37/0.39 | 0.32/0.39/0.44 |
| Text2Pos [18] | 0.14/0.25/0.31 | 0.36/0.55/0.61 | 0.48/0.68/0.74 | 0.13/0.20/0.30 | 0.33/0.42/0.49 | 0.43/0.61/0.65 |
| RET [19] | 0.19/0.30/0.37 | 0.44/0.62/0.67 | 0.52/0.72/0.78 | 0.16/0.25/0.29 | 0.35/0.51/0.56 | 0.46/0.65/0.71 |
| Text2Loc [4] | 0.37/0.57/0.63 | 0.68/0.85/0.87 | 0.77/0.91/0.93 | 0.33/0.48/0.52 | 0.60/0.75/0.78 | 0.70/0.84/0.86 |
| Des4Pos (Our) | **0.45/0.63/0.69** | **0.76/0.89/0.92** | **0.84/0.94/0.96** | **0.40/0.54/0.57** | **0.68/0.80/0.82** | **0.77/0.87/0.89** |
| *MNCL [20]* | *0.52/0.73/0.78* | *0.79/0.94/0.95* | *0.85/0.96/0.97* | *0.46/0.65/0.68* | *0.73/0.87/0.89* | *0.80/0.92/0.93* |
| *MNCL+Des4Pos* | ***0.53/0.75/0.80*** | ***0.81/0.95/0.96*** | ***0.86/0.97/0.98*** | ***0.48/0.67/0.71*** | ***0.74/0.89/0.90*** | ***0.81/0.94/0.95*** |

TABLE II
TEXT-TO-POINT-CLOUD-SUBMAP RETRIEVAL ACCURACY OF DES4POS AND SOTA TWO-STAGE METHODS IN COARSE STAGE ON THE KITTI360POSE DATASET. MNCL+DES4POS SIGNIFIES THE INTEGRATION OF THE TEXT ENCODER (STE), AND THE POINT CLOUD ENCODERS (MFAM AND LSTM) INTO THE MNCL ARCHITECTURE.

| Methods | Submap Retrieval Recall ↑ | | | | | |
|---|---|---|---|---|---|---|
| | Validation Set | | | Test Set | | |
| | k = 1 | k = 3 | k = 5 | k = 1 | k = 3 | k = 5 |
| Text2Pos | 0.14 | 0.28 | 0.37 | 0.12 | 0.25 | 0.33 |
| RET | 0.18 | 0.34 | 0.44 | 0.15 | 0.29 | 0.37 |
| Text2Loc | 0.31 | 0.54 | 0.64 | 0.28 | 0.49 | 0.58 |
| Des4Pos (Our) | **0.37** | **0.63** | **0.74** | **0.33** | **0.54** | **0.64** |
| *MNCL* | *0.48* | *0.72* | *0.81* | *0.43* | *0.66* | *0.75* |
| ***MNCL+Des4Pos*** | ***0.50*** | ***0.76*** | ***0.84*** | ***0.45*** | ***0.69*** | ***0.77*** |

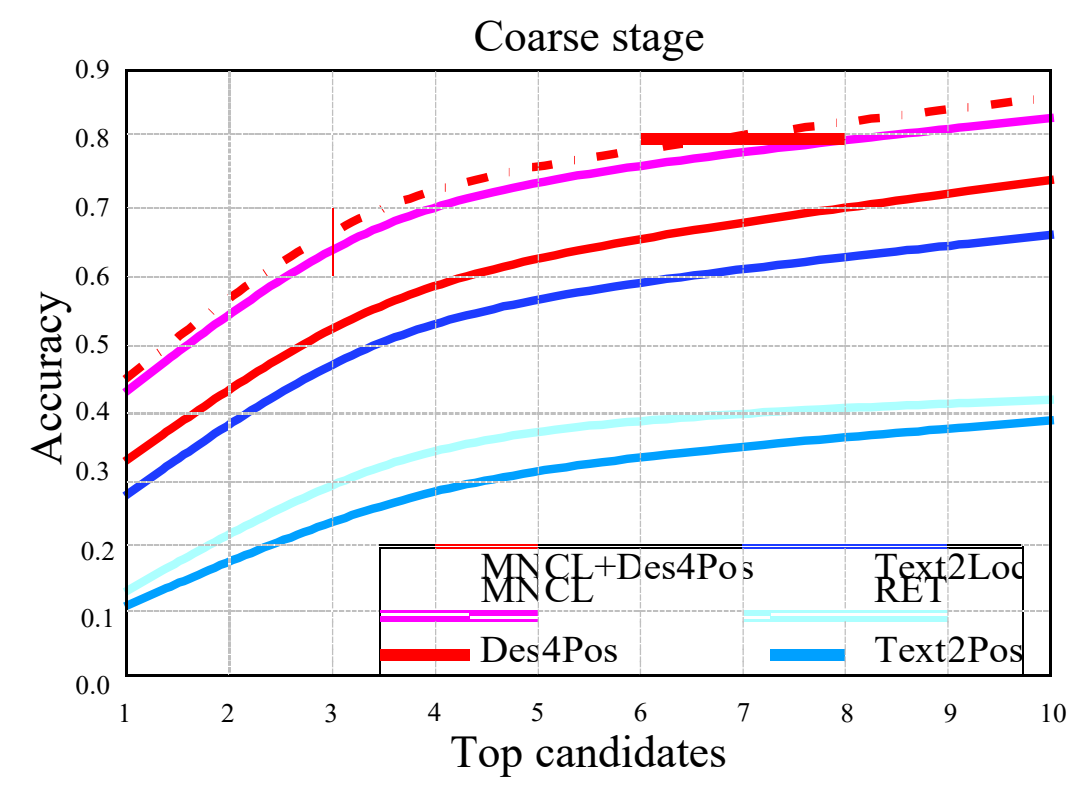

Fig. 4. The performance of text-to-submap retrieval in the coarse stage on KITTI360Pose dataset.

To further validate the effectiveness of our method, we conducted a comparative study with the *concurrent* state-of-the-art method, MNCL. While Des4Pos's primary innovation is its cross-modal spatial-enhanced encoder, MNCL's core contribution is its novel loss function. Specifically, MNCL learns multiple mismatched factors by minimizing the similarity of different locations at different levels. Although the original Des4Pos accuracy is lower than that of MNCL, we posit that the two methods are complementary in their fundamental approach. To demonstrate this synergy, we replaced key components in the MNCL architecture with those from Des4Pos: (1)We use the Des4Pos Text Encoder (STE) to replace MNCL's text encoder (Attention). (2)We use the Des4Pos Point Cloud Encoders (MFAM and LSTM) to replace MNCL's point cloud encoder (Attention). (3)We replace MNCL's Cascaded Attention with the Des4Pos CRA module for multi-modal fusion in the fine stage.

As shown in Table I, the Des4Pos-enhanced MNCL achieves a significant improvement in Top-1 5m/10m/15m accuracy on the test set, with gains of **2%**, **2%**, and **3%**, respectively. Moreover, it outperforms the original MNCL across all distance thresholds. This result not only substantially raises the current SOTA performance but also proves the robustness of the Des4Pos spatial-enhanced cross-modal encoder. Specifically, the encoder demonstrates excellent performance even when integrated with a different loss alignment mechanism.

*2) Performance Comparison on Coarse Stage:* Des4Pos follows a coarse-to-fine localization pipeline. To comprehensively evaluate model performance, we compare our method with existing two-stage approaches on text-to-point-cloud-submap retrieval accuracy during the coarse stage. As shown in Table II and Fig.4, our method achieves improvements of **6%**, **9%**, and **10%** in Top-1/3/5 accuracy over the previous SOTA two-stage method Text2Loc on the validation set, with corresponding gains of **5%**, **5%**, and **6%** on the test set. Moreover, after integrating the Des4Pos's cross-modal spatial enhancement encoder into MNCL, the performance improved by **2**%, **4**%, and **3**% in Top-1, Top-3, and Top-5 accuracy on the validation set over the original MNCL, respectively. The gains on the test set were **2**%, **3**%, and **2**% for the same metrics. This improvement strongly validates the significant advantages of Des4Pos in coarse stage and the effectiveness of its module design.

The performance improvement primarily stems from two core designs. Our MFAM thoroughly encodes local geometric information for point-cloud processing while employing a bidirectional LSTM to capture global spatial relationships between object instances. The STE effectively mitigates the semantic gap between text and point-cloud modalities for text processing. The synergistic operation of these modules enables efficient cross-modal feature alignment in the coarse stage.

*3) Performance Comparison on Fine Stage:* We further compare the performance of different two-stage methods in the fine stage, as shown in Fig. 5. When the coarse stage model is fixed, the fine stage results show that Des4Pos consistently outperforms existing methods under varying experimental conditions. These improvements further validate the effectiveness of the CRA module in preserving modality-specific feature differences through residual connections during cross-modal fusion.

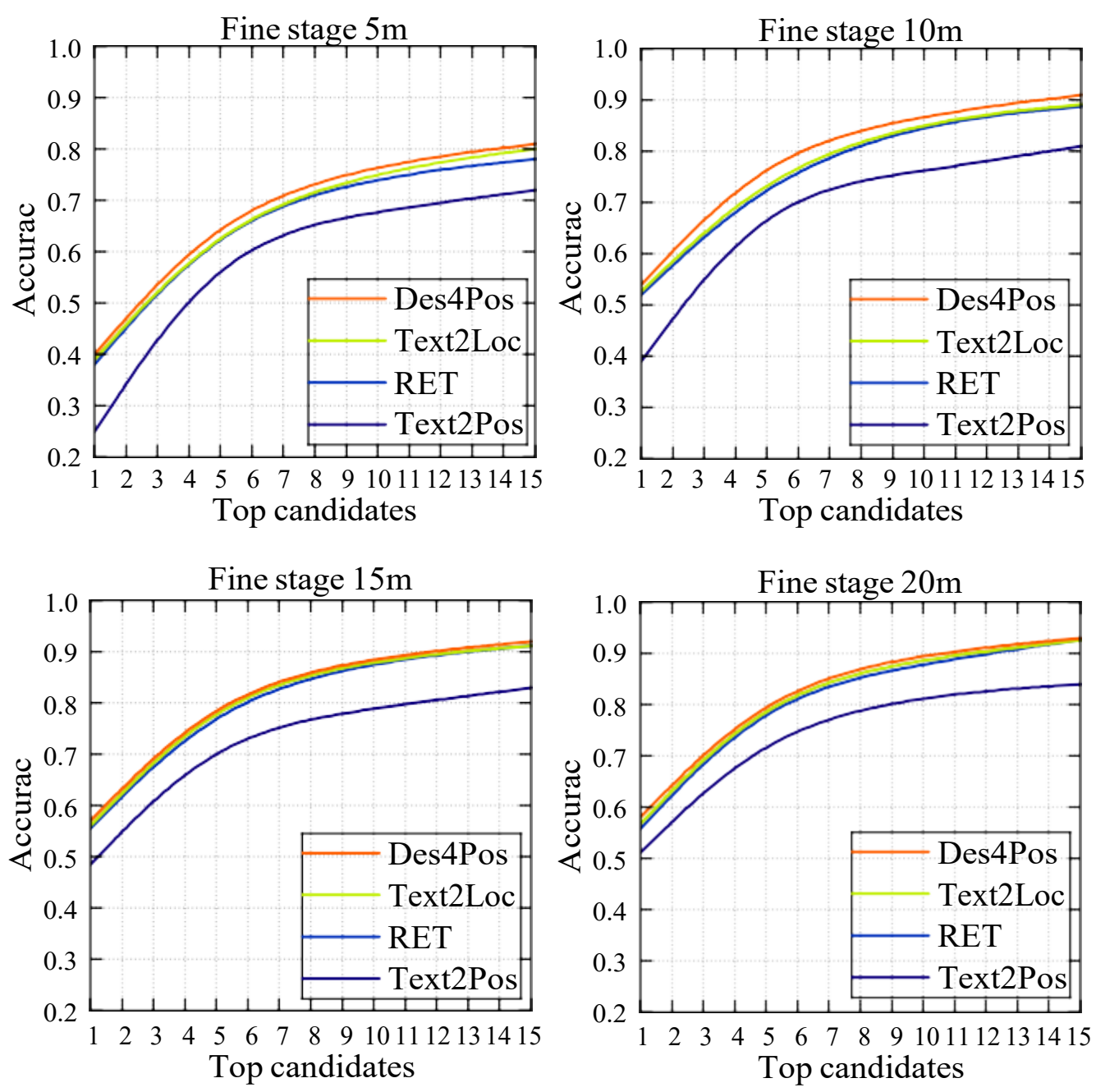

Fig. 5. We compare the fine stage performance of different two-stage methods on the KITTI360Pose test set. To ensure a fair evaluation, all methods use the coarse-stage retrieval results generated by Des4Pos.

### *C. Ablation Study*

This section validates the effectiveness of key modules in Des4Pos through systematic ablation studies.

*1) Ablation Study for Point-Cloud Encoder:* As shown in the first row in Fig. 6, we sequentially remove the MFAM and the bidirectional LSTM from the point-cloud encoder.

The experimental results demonstrate that removing the MFAM module leads to **13%**, **7%**, and **11%** declines in Top1/5/10 accuracy within 5 meters, respectively. These results confirm the efficacy of MFAM in multi-scale local feature aggregation: MFAM first captures local geometric features at varying scales, then fuses cross-scale features through dynamic attention-weighted interaction, thereby significantly improving the discriminability of local descriptors.

Furthermore, removing the bidirectional LSTM module reduces the Top1/5/10 accuracy within 5 meters by **11%**, **5%**, and **10%**, respectively. This indicates the critical role of bidirectional LSTM in global feature representations. By performing bidirectional scanning, the module effectively captures long-range spatial dependencies among point-cloud instances. Additionally, the gated mechanism within the LSTM filters key features and suppresses redundant information, ultimately generating discriminative global descriptors. Experimental results indicate that bidirectional LSTM and MFAM complement each other. Their collaboration establishes a hierarchical point-cloud feature representation framework, significantly improving the performance of Des4Pos.

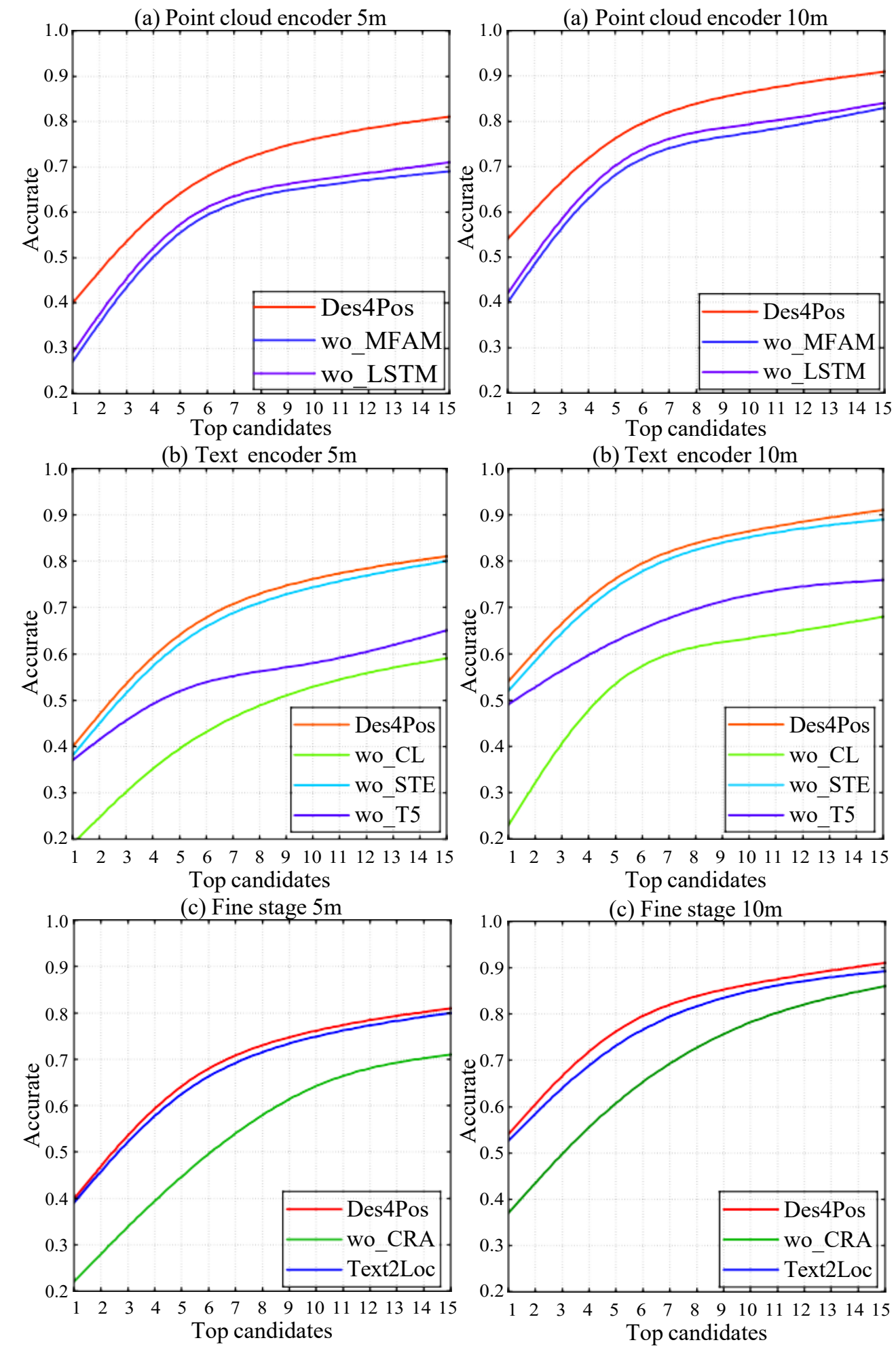

Fig. 6. We divide Des4Pos into three parts: (a) the point-cloud encoder, (b) the text encoder, and (c) the fine stage. For each part, we conduct comprehensive ablation experiments on the KITTI360Pose dataset.

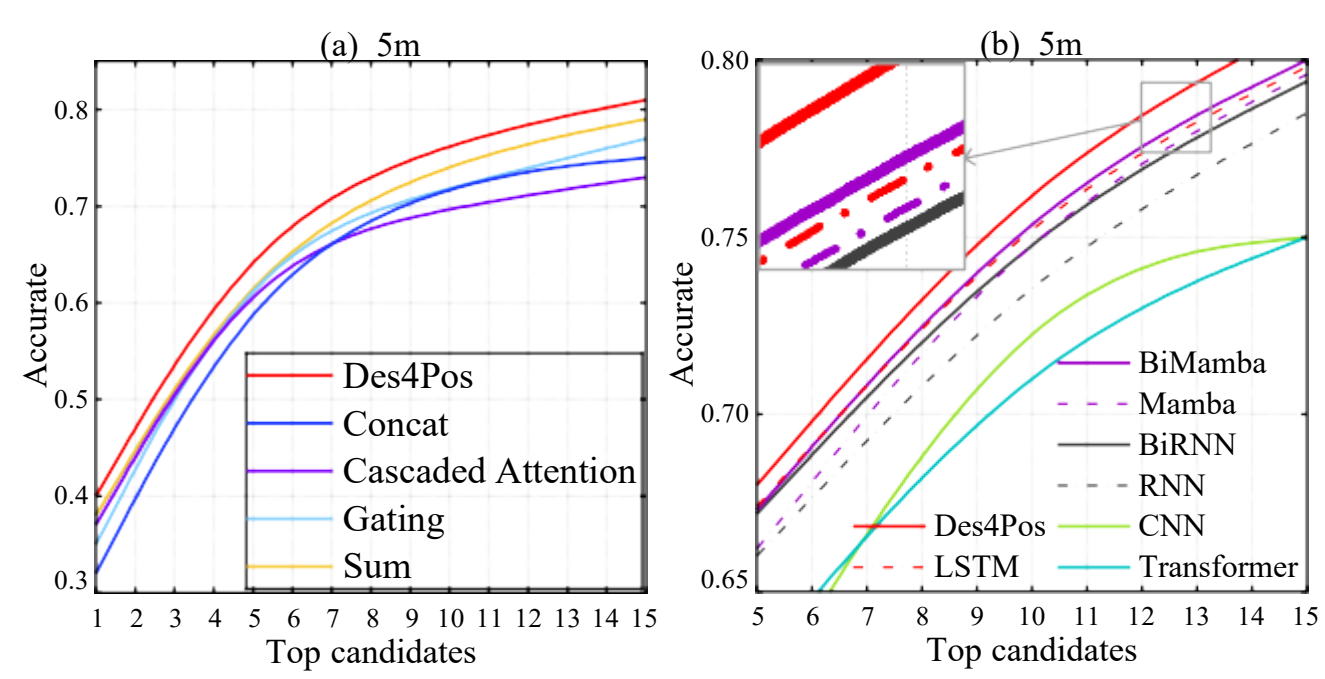

Fig. 7. In (a), we compare different methods for fusing multi-scale features within the MFAM. In (b), we compare different methods for enhancing long-range object relationships after the MFAM.

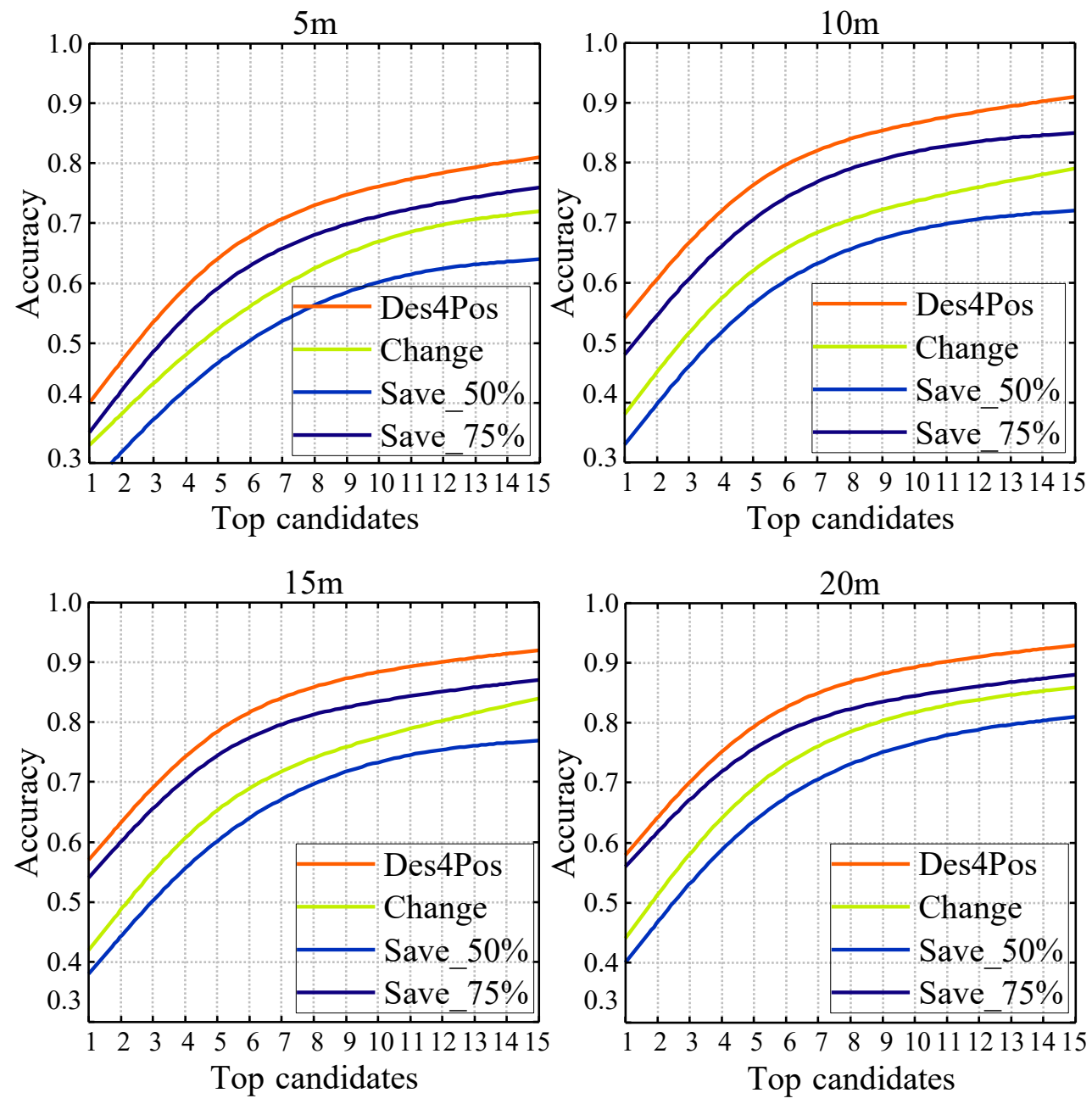

Fig. 8. We conduct a robustness analysis of Des4Pos on the KITTI360Pose test set. The experimental setup includes three testing scenarios: Save%_75 denotes retaining 75% of the sentences in each text description group during testing. Save%_50 denotes retaining 50% of the sentences in each text description group. 'Change' indicates randomly replacing one correct description with an incorrect one within the description group.

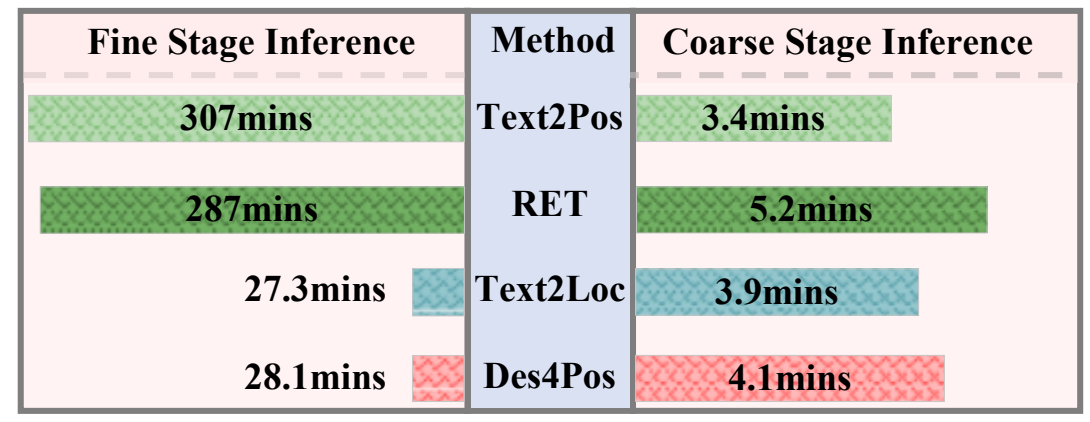

Fig. 9. We conduct a computational analysis for both the coarse and fine stages during inference, measuring the total time (in minutes) required to process the entire KITTI360Pose test set. This process involves finding 11,505 places based on text queries.

*2) C:* omparative Experiments for MFAM and LSTM To further explore the effectiveness of MFAM, we compared different methods for integrating features at different scales, as shown in Fig. 7 (a). Sum fuses features by adding the feature maps element-wise. Concat fuses features by concatenating them directly before a linear transformation. Gating fuses features by employing a mechanism to selectively combine them from different scales. Cascaded Attention fuses features using a two-stage attention process: first fusing features from the 1×1 and 3×3 convolutions, and then integrating the 5×5 features with a second attention mechanism. Our attention-weight-based fusion strategy used in the MFAM surpasses these traditional multi-scale attention fusion methods. Our method places a greater emphasis on the commonalities within the multi-scale features, which strengthens the spatial positional relationships. This makes it superior to the other fusion strategies we evaluated.

We conducted further experiments to evaluate the impact of using a bidirectional LSTM on our model's performance, as shown in Fig. 7 (b). We replaced the bidirectional LSTM with several alternative architectures, including a unidirectional LSTM, Mamba, Bidirectional Mamba (BiMamba), RNN, Bidirectional RNN (BiRNN), CNN, and Transformer.

The results clearly show that sequential methods (LSTM, Mamba [47], RNN) consistently outperform non-sequential methods (Transformer, CNN). Within the sequential approaches, the bidirectional mode yields superior performance. This is because sequential methods are better suited to processing object instances in the point cloud one by one, which strengthens their spatial positional relationships. The bidirectional mode further enhances this by incorporating complementary global information. These findings provide stronger evidence for the effectiveness of our design to model global spatial relationships using a sequential approach.

*3) Ablation Study for Text Encoder:* As shown in Fig. 6 (b), we conduct systematic ablation studies for the text encoder: wo_T5 denotes replacing T5 with CLIP as the text encoder backbone; wo_STE indicates directly aligning text and point-cloud features using all encoder layers during training without introducing cross-modal prior knowledge; and wo_CL represents substituting the contrastive loss with pair loss for cross-modal alignment.

The experimental results demonstrate that replacing T5 with CLIP (wo_T5) leads to significant performance degradation, which stems from CLIP's limitations in processing long text sequences.

When we remove cross-modal prior transfer in wo_STE, a performance drop is observed. This validates the effectiveness of the two-stage transfer strategy in the STE module. In the first stage, vision-language prior knowledge is obtained from CLIP. This prior knowledge is transferred to text-to-point-cloud alignment tasks. As a result, the STE module reduces the learning difficulty of text-to-point-cloud alignment by leveraging pre-trained cross-modal knowledge.

Finally, replacing contrastive loss with pair loss (wo_CL) causes a sharp performance decline. Contrastive learning pulls matched text-point-cloud pairs closer and pushes unmatched ones apart. This process facilitates more discriminative cross-modal mappings in high-dimensional feature spaces, ultimately improving the performance of Des4Pos.

*4) Ablation Study for Fine Stage:* We conduct ablation studies on the fine stage of Des4Pos. In Fig. 6 (c), wo_CRA denotes directly concatenating point-cloud and text features instead of CRA followed by MLP for relative position offset prediction. wo_Residual indicates removing the residual connection in the CRA module.

Experimental results demonstrate that removing the CRA module leads to significant performance degradation, validating the effectiveness of the cross-modal attention fusion mechanism. Further ablation analysis on the CRA module reveals that eliminating residual connections (wo_Residual)

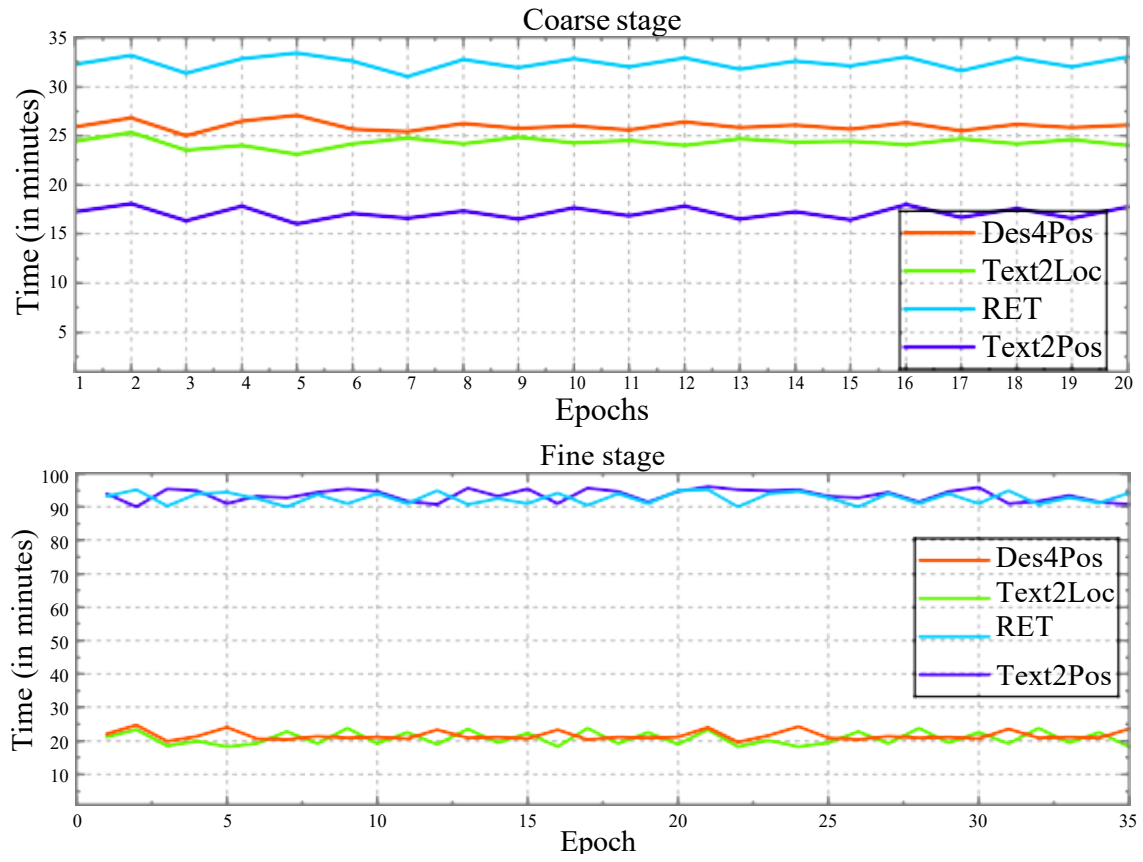

Fig. 10. We conduct a computational analysis of the coarse and fine stages during training, recording the time cost for each epoch. The coarse stage comprises 20 epochs, and the fine stage comprises 35 epochs.

reduces the top-1 localization accuracy within 5 meters by **1%** on the test set.

The CRA module uses cascaded cross-modal attention layers to deeply fuse point clouds and text features. Meanwhile, residual connections in the module preserve each modality's unique features during fusion. These features also maintain cross-modal disparity, which helps predict position offsets. This balance between fusing modalities and preserving original features significantly improves the precise localization accuracy.

### *D. Robustness Analysis*

This section conducts a robustness analysis of Des4Pos from two perspectives: descriptive sufficiency and descriptive accuracy. First, we evaluate the impact of descriptive sufficiency on Des4Pos performance by reducing the number of sentences in textual descriptions. As shown in Fig. 8, when retaining 75% of sentences as text queries, the top-1 accuracy of Des4Pos decreases by 5%, 6%, 3%, and 2% at ranges of 5m, 10m, 15m, and 20m, respectively. The positive correlation between the number of sentences and localization accuracy indicates that complete descriptions offer clearer geographical references. Furthermore, when reducing descriptive information to 50%, the system exhibits significant accuracy drops of 14%, 21%, 19%, and 18% within the same ranges. The performance drop stems from insufficient information: 75% of descriptions retain enough detail for accurate localization, but 50% of descriptions suffer severe semantic loss, leading to ambiguous matches in point-cloud maps. Second, in the descriptive accuracy test, we simulate erroneous descriptions by randomly replacing one sentence in the text queries. Experimental results demonstrate that Des4Pos achieves top-1, top-5, top-10, and top-15 accuracy of 33%, 54%, 69%, and 72%, respectively, within the 5 m range. These outcomes confirm Des4Pos' capability to maintain reliable localization performance even with partially erroneous descriptions.

### *E. Computational Resource Consumption*

We perform a computational analysis of the Des4Pos for both the interface and training phases. As shown in Fig. 9, during the interface phase, Des4Pos requires slightly more time than Text2Loc in the coarse stage. In the fine stage, Des4Pos is as efficient as Text2Loc, as both methods utilize a strategy that avoids instance matching.

Subsequently, we focus on the training phase for computational analysis, as shown in Fig. 10. In the coarse stage, Des4Pos requires approximately 2 minutes longer than Text2Loc per epoch. In the fine stage, Des4Pos's computational cost is virtually equal to that of Text2Loc. These results are consistent with the computational performance in the interface phase.

The experiments above demonstrate that our proposed Des4Pos achieves significant performance improvements under the same computational resource constraints, proving the effectiveness of the Des4Pos.

### *F. Visualization Results*

In the visualization analysis, we first compare the text-retrieved point cloud with the real point cloud of the location. Specifically, as illustrated in Fig. 11, key point matching is performed between the two point-cloud maps. These experiments demonstrate a significant number of accurately matched key points between them. Notably, these results enhance the interpretability of Des4Pos: text descriptions can represent key points in point cloud, and we use the implicit geometric and spatial information from these key points to locate positions in large-scale point cloud maps. Furthermore, these results validate the effectiveness of the Des4Pos method.

Through additional visualization analysis, we identify a set of locations retrieved from text queries, as shown in Fig. 12. For the majority of text queries, the Des4Pos method effectively retrieves multiple adjacent locations, reflecting its strong performance. However, in specific error cases such as (d). TOP 1 and TOP 2, we observe that the text descriptions exhibit partial similarities with the incorrectly localized areas. This suggests that the insufficient information in the text descriptions leads Des4Pos to incorrectly localize geographically distant but visually similar locations, which constitutes the primary reason for incorrect localization.

## V. Conclusion

We present Des4Pos, a novel two-stage vehicle-scene interaction-based localization framework designed for urban environments. In the coarse stage, our method employs an innovative Multi-scale Fusion Attention Mechanism (MFAM) in conjunction with a bidirectional LSTM to encode point clouds, capturing local features and global spatial relationships. Additionally, we propose a Stepped Text Encoder (STE) that effectively utilizes cross-modal priors to facilitate text- to-point-cloud alignment. In the fine stage, the Cascaded Residual Attention (CRA) refines positional coordinates by integrating text and point cloud features while preserving modality-specific details through residual connections, ensuring precise localization. Extensive experiments demonstrate that Des4Pos achieves a top-1 accuracy of **68%** within 5 meters, outperforming the most effective existing method Text2Loc by **8%** under the same computational constraints. Des4Pos demonstrates excellent performance across different optimization objectives. Notably, when combined with the concurrent method MNCL, it achieves a significant performance improvement. We anticipate that Des4Pos will serve as a foundational baseline, advancing research in text-based localization and inspiring future explorations in cross-modal navigation systems.

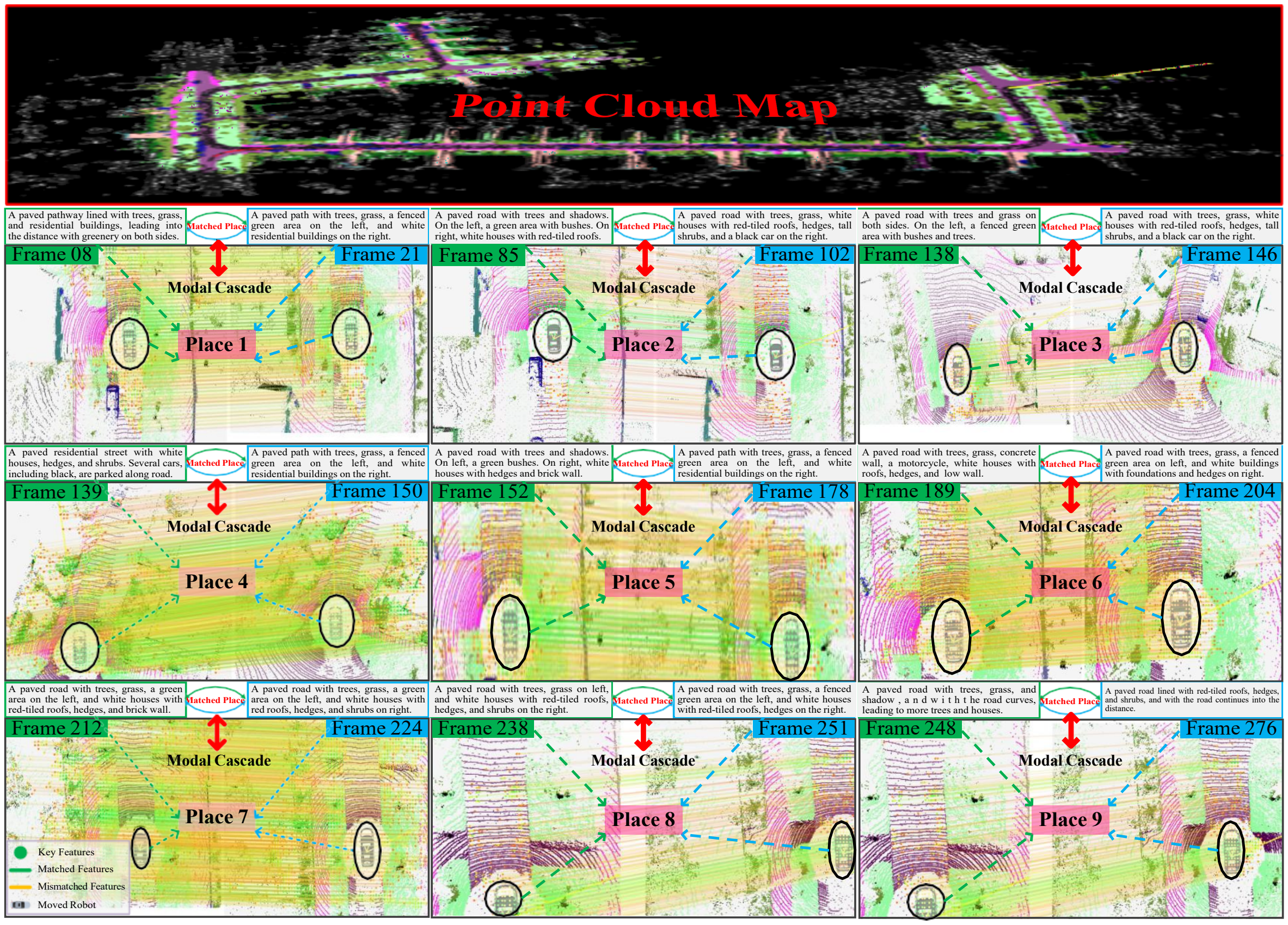

Fig. 11. In this visualization analysis experiment, Des4Pos retrieves candidate point-cloud submaps from large-scale maps based on textual descriptions, followed by keypoint matching between the retrieved submaps and the actual point-cloud observed at the target location.

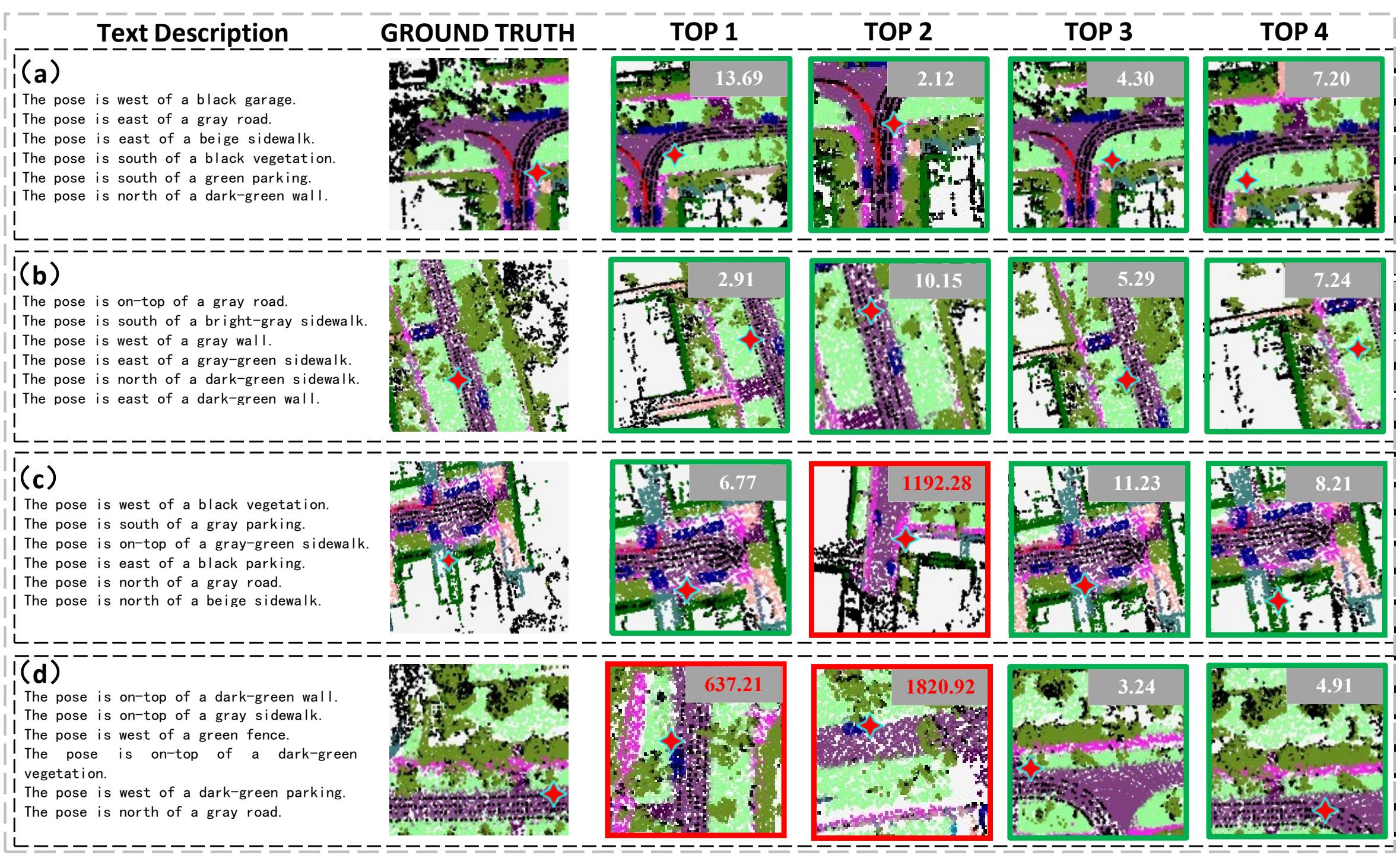

Fig. 12. In this visualization analysis experiment, we present the textual descriptions, the point clouds of the actual location (ground truth), and the top K retrieved point-cloud submaps. We mark the retrieved position in each submap with a star symbol. If the Euclidean distance between the retrieved coordinates and the actual coordinates is within 15 meters, we consider this retrieval correct and highlight the submap with a green box. Incorrect submaps are marked with red boxes. We also display the distance between each retrieved coordinate and the target coordinate in the top-right corner.

Limitations: We acknowledge some limitations. Our framework is slightly slower in both training and inference compared to previous methods, although this difference is negligible. Second, current methods, including our own, lack training and validation data for other challenging scenarios such as heavy rain, snow, and severe occlusions.

Future Research Directions: We aim to construct a larger-scale database that includes more challenging scenarios like heavy rain, severe occlusion, and non-street environments to improve the robustness and generalization of our model. We also plan to integrate large-scale models to help our framework recognize objects not present in the dataset or to use prompts to improve model robustness.

## References

[1] T. Wang, Z. Li, Y. Xu, Y. Zhai, X. Xing, K. Guo, P. Coscia, A. Genovese, V. Piuri, and F. Scotti, "Clip-vision guided few-shot metal surface defect recognition," *IEEE Transactions on Industrial Informatics*, 2025.

[2] M. Shi, Y. Su, X. Lin, B. Zao, S. Kong, and M. Tan, "Frame as video clip: Proposal-free moment retrieval by semantic aligned frames," *IEEE Transactions on Industrial Informatics*, 2024.

[3] Z. Zhang, C. Niu, Z. Zhao, X. Zhang, and X. Chen, "Small object few-shot segmentation for vision-based industrial inspection," *IEEE Transactions on Industrial Informatics*, 2025.

[4] Y. Xia, L. Shi, Z. Ding, J. F. Henriques, and D. Cremers, "Text2loc: 3d point cloud localization from natural language," in *Proceedings of the IEEE/CVF conference on computer vision and pattern recognition*, 2024, pp. 14 958–14 967.

[5] Z. Li, P. Xu, Z. Dong, R. Zhang, and Z. Deng, "Feature-level knowledge distillation for place recognition based on soft-hard labels teaching paradigm," *IEEE Transactions on Intelligent Transportation Systems*, 2024.

[6] Z. Li, T. Shang, and P. Xu, "Multi-modal attention perception for intelligent vehicle navigation using deep reinforcement learning," *IEEE Transactions on Intelligent Transportation Systems*, 2025.

[7] C. Wang, S. Chen, Y. Song, R. Xu, Z. Zhang, J. Zhang, H. Yang, Y. Zhang, K. Fu, S. Du, *et al.*, "Focus on local: Finding reliable discriminative regions for visual place recognition," in *Proceedings of the AAAI Conference on Artificial Intelligence*, vol. 39, no. 7, 2025, pp. 7536–7544.

[8] Z. Li, P. Xu, and T. Shang, "Cwpformer: Towards high-performance visual place recognition for robot with cross-weight attention learning," *IEEE Transactions on Artificial Intelligence*, 2025.

[9] F. Lu, T. Jin, X. Lan, L. Zhang, Y. Liu, Y. Wang, and C. Yuan, "Selavpr++: Towards seamless adaptation of foundation models for efficient place recognition," *arXiv preprint arXiv:2502.16601*, 2025.

[10] A. Ali-Bey, B. Chaib-Draa, and P. Giguere, "Mixvpr: Feature mixing for visual place recognition," in *Proceedings of the IEEE/CVF winter conference on applications of computer vision*, 2023, pp. 2998–3007.

[11] L. Luo, S. Zheng, Y. Li, Y. Fan, B. Yu, S.-Y. Cao, J. Li, and H.-L. Shen, "Bevplace: Learning lidar-based place recognition using bird's eye view images," in *Proceedings of the IEEE/CVF International Conference on Computer Vision*, 2023, pp. 8700–8709.

[12] L. Hui, H. Yang, M. Cheng, J. Xie, and J. Yang, "Pyramid point cloud transformer for large-scale place recognition," in *Proceedings of the IEEE/CVF International Conference on Computer Vision*, 2021, pp. 6098–6107.

[13] C. Masone and B. Caputo, "A survey on deep visual place recognition," *IEEE Access*, vol. 9, pp. 19 516–19 547, 2021.

[14] P. Yin, J. Jiao, S. Zhao, L. Xu, G. Huang, H. Choset, S. Scherer, and J. Han, "General place recognition survey: Towards real-world autonomy," *IEEE Transactions on Robotics*, 2025.

[15] Z. Du, S. Ji, and K. Khoshelham, "3-d lidar-based place recognition techniques: A review of the past ten years," *IEEE Transactions on Instrumentation and Measurement*, vol. 73, pp. 1–24, 2024.

[16] Y. Zhang, P. Shi, and J. Li, "Lidar-based place recognition for autonomous driving: A survey," *ACM Computing Surveys*, vol. 57, no. 4, pp. 1–36, 2024.

[17] Z. Li, T. Shang, P. Xu, and Z. Deng, "Place recognition meet multiple modalities: a comprehensive review, current challenges and future development," *Artificial Intelligence Review*, vol. 58, no. 11, p. 363, 2025.

[18] M. Kolmet, Q. Zhou, A. Ošep, and L. Leal-Taixé, "Text2pos: Text-to-point-cloud cross-modal localization," in *Proceedings of the IEEE/CVF Conference on Computer Vision and Pattern Recognition*, 2022, pp. 6687–6696.

[19] G. Wang, H. Fan, and M. Kankanhalli, "Text to point cloud localization with relation-enhanced transformer," in *Proceedings of the AAAI Conference on Artificial Intelligence*, vol. 37, no. 2, 2023, pp. 2501–2509.

[20] D. Liu, S. Huang, W. Li, S. Shen, and C. Wang, "Text to point cloud localization with multi-level negative contrastive learning," in *Proceedings of the AAAI Conference on Artificial Intelligence*, vol. 39, no. 5, 2025, pp. 5397–5405.

[21] Y. Liao, J. Xie, and A. Geiger, "Kitti-360: A novel dataset and benchmarks for urban scene understanding in 2d and 3d," *IEEE Transactions on Pattern Analysis and Machine Intelligence*, vol. 45, no. 3, pp. 3292–3310, 2022.

[22] M. A. Uy and G. H. Lee, "Pointnetvlad: Deep point cloud based retrieval for large-scale place recognition," in *Proceedings of the IEEE conference on computer vision and pattern recognition*, 2018, pp. 4470–4479.

[23] C. R. Qi, H. Su, K. Mo, and L. J. Guibas, "Pointnet: Deep learning on point sets for 3d classification and segmentation," in *Proceedings of the IEEE conference on computer vision and pattern recognition*, 2017, pp. 652–660.

[24] R. Arandjelovic, P. Gronat, A. Torii, T. Pajdla, and J. Sivic, "Netvlad: Cnn architecture for weakly supervised place recognition," in *Proceedings of the IEEE conference on computer vision and pattern recognition*, 2016, pp. 5297–5307.

[25] S. Zheng, Y. Li, Z. Yu, B. Yu, S.-Y. Cao, M. Wang, J. Xu, R. Ai, W. Gu, L. Luo, *et al.*, "I2p-rec: Recognizing images on large-scale point cloud maps through bird's eye view projections," in *2023 IEEE/RSJ International Conference on Intelligent Robots and Systems (IROS)*. IEEE, 2023, pp. 1395–1400.

[26] F. Liu, Y. Cao, X. Cheng, J. Wu, W. Gu, and L. Liu, "Confidence factor based robust localization algorithm with visual-inertial-lidar fusion in underground space," *IEEE Transactions on Circuits and Systems for Video Technology*, 2025.

[27] Z. Liu, S. Zhou, C. Suo, P. Yin, W. Chen, H. Wang, H. Li, and Y.-H. Liu, "Lpd-net: 3d point cloud learning for large-scale place recognition and environment analysis," in *Proceedings of the IEEE/CVF international conference on computer vision*, 2019, pp. 2831–2840.

[28] J. Du, R. Wang, and D. Cremers, "Dh3d: Deep hierarchical 3d descriptors for robust large-scale 6dof relocalization," in *Computer Vision–ECCV 2020: 16th European Conference, Glasgow, UK, August 23–28, 2020, Proceedings, Part IV 16*. Springer, 2020, pp. 744–762.

[29] Y. Xia, Y. Xu, S. Li, R. Wang, J. Du, D. Cremers, and U. Stilla, "Soe-net: A self-attention and orientation encoding network for point cloud based place recognition," in *Proceedings of the IEEE/CVF Conference on computer vision and pattern recognition*, 2021, pp. 11 348–11 357.

[30] J. Knights, P. Moghadam, M. Ramezani, S. Sridharan, and C. Fookes, "Incloud: Incremental learning for point cloud place recognition," in *2022 IEEE/RSJ International Conference on Intelligent Robots and Systems (IROS)*. IEEE, 2022, pp. 8559–8566.

[31] T.-X. Xu, Y.-C. Guo, Z. Li, G. Yu, Y.-K. Lai, and S.-H. Zhang, "Transloc3d: Point cloud based large-scale place recognition using adaptive receptive fields," *arXiv preprint arXiv:2105.11605*, 2021.

[32] Z. Zhou, C. Zhao, D. Adolfsson, S. Su, Y. Gao, T. Duckett, and L. Sun, "Ndt-transformer: Large-scale 3d point cloud localisation using the normal distribution transform representation," in *2021 IEEE international conference on robotics and automation (ICRA)*. IEEE, 2021, pp. 5654–5660.

[33] J. Komorowski, "Minkloc3d: Point cloud based large-scale place recognition," in *Proceedings of the IEEE/CVF winter conference on applications of computer vision*, 2021, pp. 1790–1799.

[34] D. Cattaneo, M. Vaghi, and A. Valada, "Lcdnet: Deep loop closure detection and point cloud registration for lidar slam," *IEEE Transactions on Robotics*, vol. 38, no. 4, pp. 2074–2093, 2022.

[35] J. Komorowski, M. Wysoczanska, and T. Trzcinski, "Egonn: Egocentric neural network for point cloud based 6dof relocalization at the city scale," *IEEE Robotics and Automation Letters*, vol. 7, no. 2, pp. 722–729, 2021.

[36] K. Vidanapathirana, M. Ramezani, P. Moghadam, S. Sridharan, and C. Fookes, "Logg3d-net: Locally guided global descriptor learning for 3d place recognition," in *2022 International Conference on Robotics and Automation (ICRA)*. IEEE, 2022, pp. 2215–2221.

[37] L. Zhao, K.-K. Ma, Z. Liu, Q. Yin, and J. Chen, "Real-time scene-aware lidar point cloud compression using semantic prior representation," *IEEE Transactions on Circuits and Systems for Video Technology*, vol. 32, no. 8, pp. 5623–5637, 2022.

[38] C. Raffel, N. Shazeer, A. Roberts, K. Lee, S. Narang, M. Matena, Y. Zhou, W. Li, and P. J. Liu, "Exploring the limits of transfer learning with a unified text-to-text transformer," *Journal of machine learning research*, vol. 21, no. 140, pp. 1–67, 2020.

[39] D. Z. Chen, A. X. Chang, and M. Nießner, "Scanrefer: 3d object localization in rgb-d scans using natural language," in *European conference on computer vision*. Springer, 2020, pp. 202–221.

[40] P. Achlioptas, A. Abdelreheem, F. Xia, M. Elhoseiny, and L. Guibas, "Referit3d: Neural listeners for fine-grained 3d object identification in real-world scenes," in *Computer Vision–ECCV 2020: 16th European Conference, Glasgow, UK, August 23–28, 2020, Proceedings, Part I 16*. Springer, 2020, pp. 422–440.

[41] Z. Yuan, X. Yan, Y. Liao, R. Zhang, S. Wang, Z. Li, and S. Cui, "Instancerefer: Cooperative holistic understanding for visual grounding on point clouds through instance multi-level contextual referring," in *Proceedings of the IEEE/CVF International Conference on Computer Vision*, 2021, pp. 1791–1800.

[42] M. Feng, Z. Li, Q. Li, L. Zhang, X. Zhang, G. Zhu, H. Zhang, Y. Wang, and A. Mian, "Free-form description guided 3d visual graph network for object grounding in point cloud," in *Proceedings of the IEEE/CVF international conference on computer vision*, 2021, pp. 3722–3731.

[43] R. Zhang, Z. Guo, W. Zhang, K. Li, X. Miao, B. Cui, Y. Qiao, P. Gao, and H. Li, "Pointclip: Point cloud understanding by clip," in *Proceedings of the IEEE/CVF conference on computer vision and pattern recognition*, 2022, pp. 8552–8562.

[44] F. Yu, Z. Wang, D. Li, P. Zhu, X. Liang, X. Wang, and M. Okumura, "Towards cross-modal point cloud retrieval for indoor scenes," in *International Conference on Multimedia Modeling*. Springer, 2024, pp. 89–102.

[45] S. He, H. Ding, X. Jiang, and B. Wen, "Segpoint: Segment any point cloud via large language model," in *European Conference on Computer Vision*. Springer, 2024, pp. 349–367.

[46] R. Xu, X. Wang, T. Wang, Y. Chen, J. Pang, and D. Lin, "Pointllm: Empowering large language models to understand point clouds," in *European Conference on Computer Vision*. Springer, 2024, pp. 131–147.

[47] A. Gu and T. Dao, "Mamba: Linear-time sequence modeling with selective state spaces," *arXiv preprint arXiv:2312.00752*, 2023.